\documentclass[pdflatex,sn-mathphys-num]{sn-jnl}

\usepackage{graphicx}%
\usepackage{multirow}%
\usepackage{amsmath,amssymb,amsfonts}%
\usepackage{amsthm}%
\usepackage{mathrsfs}%
\usepackage[title]{appendix}%
\usepackage[table]{xcolor} 
\usepackage{textcomp}%
\usepackage{manyfoot}%
\usepackage{booktabs}%
\usepackage{algorithm}%
\usepackage{algorithmicx}%
\usepackage{algpseudocode}%
\usepackage{listings}%
\usepackage{cleveref}
\usepackage{wrapfig}
\usepackage[
    textwidth=9.5mm,
    colorinlistoftodos, 
    textsize=scriptsize, 
    disable 
]{todonotes}
\newif\ifshowannotations

\showannotationsfalse 
\newcommand{\highlight}[1]{\ifshowannotations{\color{orange}\textbf{#1}}\else{#1}\fi}

\theoremstyle{thmstyleone}%
\theoremstyle{thmstyletwo}%

\theoremstyle{thmstylethree}%

\DeclareUnicodeCharacter{2212}{-}

\begin{document}

\title[Article Title]{SHADeS: Self-supervised Monocular Depth Estimation Through Non-Lambertian Image Decomposition}


\author*{\fnm{Rema} \sur{Daher}}\email{rema.daher.20@ucl.ac.uk}
\author{\fnm{Francisco} \sur{Vasconcelos}}\email{f.vasconcelos@ucl.ac.uk;}
\author{\fnm{Danail} \sur{Stoyanov}}\email{danail.stoyanov@ucl.ac.uk}

\affil{\orgdiv{UCL Hawkes Institute, Dept. Computer Science}, \orgname{University College London}, \orgaddress{\street{Gower Street}, \city{London}, \postcode{WC1E 6BT}, \country{UK}}}


\abstract{\textbf{Purpose:} Visual 3D scene reconstruction can support colonoscopy navigation. It can help in recognising which portions of the colon have been visualised and characterising the size and shape of polyps. This is still a very challenging problem due to complex illumination variations, including abundant specular reflections. We investigate how to effectively decouple light and depth in this problem.

\textbf{Methods:} We introduce a self-supervised model that simultaneously characterises the shape and lighting of the visualised colonoscopy scene. Our model estimates shading, albedo, depth, and specularities (SHADeS) from single images. Unlike previous approaches (IID \highlight{\cite{li2024image}}\todo{R2.2}), we use a non-Lambertian model that treats specular reflections as a separate light component. The implementation of our method is available at \url{https://github.com/RemaDaher/SHADeS}.

\textbf{Results:} We demonstrate on real colonoscopy images (Hyper Kvasir) that previous models for light decomposition (IID) and depth estimation (MonoVIT, ModoDepth2) are negatively affected by specularities. In contrast, SHADeS can simultaneously produce light decomposition and depth maps that are robust to specular regions. We also perform a quantitative comparison on phantom data (C3VD) where we further demonstrate the robustness of our model.

\textbf{Conclusion:} Modelling specular reflections improves depth estimation in colonoscopy. We propose an effective self-supervised approach that uses this insight to jointly estimate light decomposition and depth. Light decomposition has the potential to help with other problems, such as place recognition within the colon.}




\maketitle

\section{Introduction} \label{sec:introduction}

Colorectal cancer is the third most common cancer worldwide with a 47\% fatality rate \cite{WCRF2022}. Early diagnosis of colorectal cancer plays a key role in improving survival rates \cite{WHO2023}. However, only 40\% of colorectal cancers are detected early on \cite{DCC}. One main reason is the difficult visibility conditions in colonoscopy. Computer vision can assist surgeons with visibility through 3D reconstruction, navigation, and polyp detection. In particular, 3D reconstruction could aid in identifying missed regions, characterising polyps, comparing screenings, training endoscopists, and autonomous navigation.

We focus on monocular depth estimation, a crucial part of endoscopic 3D reconstruction and navigation. State-of-the-art (SOTA) methods such as MonoViT \cite{zhao2022monovit} have achieved impressive results on non-medical images. However, they still struggle with visibility challenges in endoscopy such as light variations and reflections due to the close-range scene with frequent motion blur and sub-optimal focus. While some non-learning methods \cite{liu2024monocular} have been proposed to tackle this problem, deep learning is still the predominant approach in recent research. Deep networks can be trained either in a supervised manner using virtual or phantom simulated data, or in a self-supervised manner with real endoscopy data. Self-supervised approaches are currently the SOTA in monocular depth estimation for endoscopy, yet they still struggle with challenging illumination conditions, such as abundant specular reflections in endoscopic images. 

\begin{wrapfigure}{l}{0.5\textwidth}
    \centering
    \includegraphics[width=0.48\textwidth]{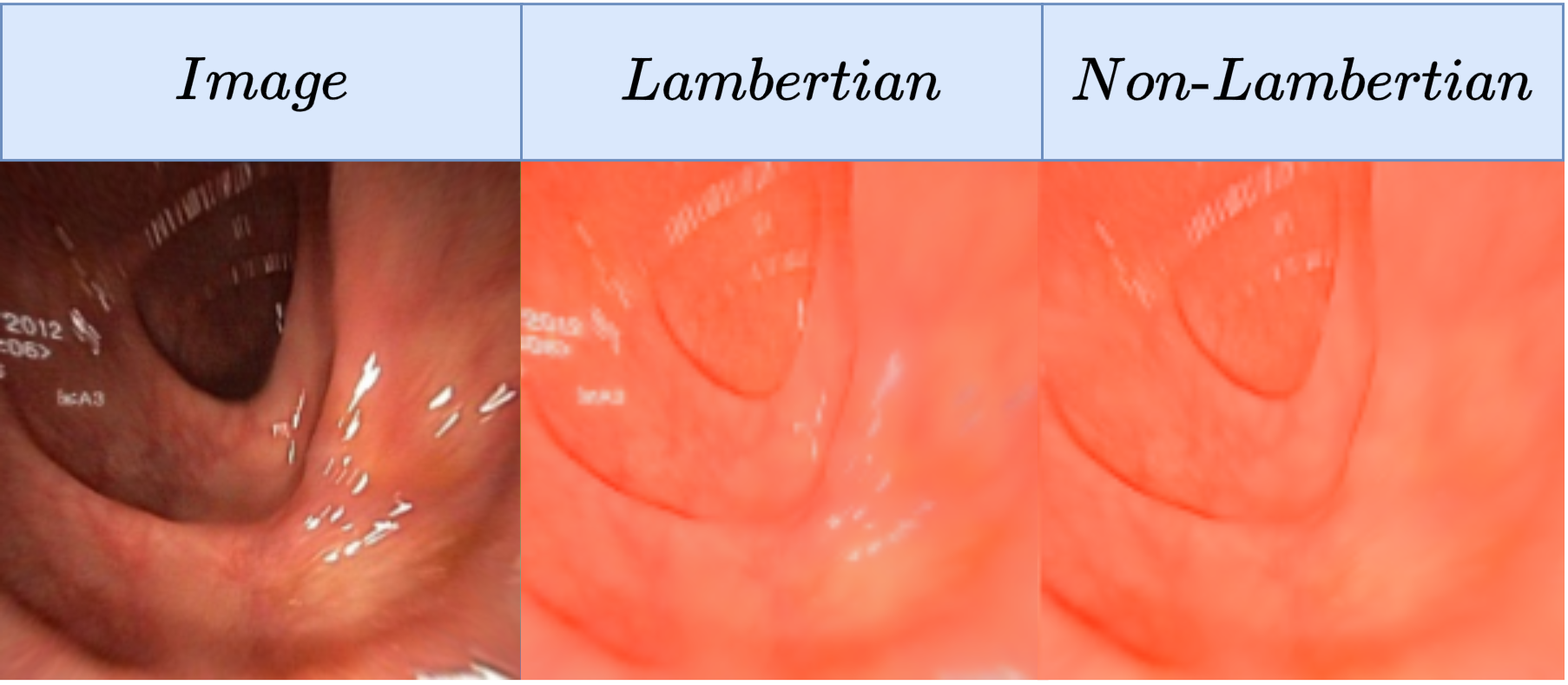}
    \caption{Extracted albedo from a Lambertian (IID) versus our non-Lambertian model (SHADeS). The specular reflections produce significantly fewer artefacts with our model.}
    \label{fig:motivation}
\end{wrapfigure}

This paper proposes a novel self-supervised approach that jointly estimates depth and decomposes an image into different light components. We take as inspiration the model in \cite{li2024image} \highlight{(IID; short for IID-SfMLearner)}\todo{R2.2}, which simultaneously estimates depth and decomposes the image ($I$) into albedo ($A$) and shading ($S$) following a Lambertian model assumption ($I = AS$). In contrast, we consider specular masks ($M$) as a third image component, following the relation $I=AS+M$. We do so because handling specular highlights has improved many computer vision tasks \cite{daher2023temporal,daher2023cyclesttn}. \Cref{fig:motivation} shows that the Lambertian model cannot distinguish between specular reflections and the underlying albedo ($A$), while our model can extract the albedo free of artefacts. Beyond the raw outputs of our model (depth, light components), it can also implicitly perform semantic segmentation of specular reflections by binarising $M$ as well as specularity removal through image inpainting ($I - M = AS$). In summary, our contributions are as follows: 
\begin{enumerate} 
    \item We propose a novel self-supervised monocular depth estimation framework that is more robust to specular reflections than the SOTA (IID, Monodepth2, MonoViT) as demonstrated on real (Hyper Kvasir) and phantom colon data (C3VD).
    \item Our model jointly estimates depth, albedo, shading, and specular reflections. This is a direct upgrade from IID, which only estimates depth, shading and albedo. We demonstrate that our model can effectively decouple albedo from specular reflections, while IID extracts albedo with specular artefacts.
    \item We can combine the different outputs of our model to implicitly estimate specularity segmentation masks as well as inpainted images without specular reflections.
\end{enumerate}


\section{Related Work}\label{sec:lit}

In recent years, monocular depth estimation has been dominated both in terms of popularity and performance by self-supervised approaches, and therefore we focus this section on these. SfMLearner \cite{zhou2017unsupervised} was one of the pioneering methods of this kind. It introduced the popular concept of jointly training depth and camera pose regression networks using a loss that measures re-projected photometric consistency on pairs of overlapping views. Most of the more recent self-supervised approaches all follow a similar training methodology. Monodepth2 \cite{godard2019digging} adds a multi-scale appearance matching loss to address occluded pixels as well as an auto-masking technique to ignore static pixels that generate infinity depth values. In parallel, SC-SfMLearner \cite{bian2019unsupervised} introduced a constraint for scale consistency and added a self-discovered mask to address dynamic scenes and occlusions. Many works have built upon these methods, with MonoViT \cite{zhao2022monovit} being a notable example with state-of-the-art performance that adopts the Monodepth2 methodology while using a transformer-based depth network. 

However, these methods have sub-optimal performance when applied to endoscopy data. One of the reasons is that they all assume the visualised scene is approximately a Lambertian surface, i.e. any 3D location is viewed with the same colour and light intensity from any viewpoint. However, in endoscopy, this is not true due to the moving light source and the visualised wet tissue being highly reflective and deformable. 

In the endoscopic domain, some methods have incorporated model-free learning based models to estimate an offset that compensates small light changes in different viewpoints. One of the first solutions of this kind proposed a linear affine brightness transformer that was added to the photometric loss \cite{ozyoruk2021endoslam}. This was extended in \cite{rau2023task} by applying domain adaptation so that both real and synthetic data can be combined during training.  To further incorporate the appearance changes in endoscopy, AF-SfMLearner \cite{shao2022self} added appearance flow and correspondence networks. In \cite{zhou2023tackling}, a conﬁdence-based colour offset penalty is added to the appearance flow network to improve low-texture and drastic illumination ﬂuctuations. Some have also introduced temporal information to AF-SfMLearner \cite{lou2024ws,shi2024long}.

A different type of methods attempt to filter out regions likely to be inconsistent, such as specular reflections. This can be achieved with a separate specularity detection algorithm that either masks out regions during loss computation \cite{li2023endodepthl,yue2023tcl}, or is utilised to learn how to reconstruct surface texture underneath specular regions \cite{wu2023unleashing}. In \cite{liao2024self}, a multitask PoseNet is incorporated to generate pose and two types of masks: one for photometric loss focused on specularities and another for geometric consistency loss focused on deformations. In \cite{rodriguez2022uncertain}, specular highlights are implicitly incorporated by minimizing uncertainty estimated through Bayesian or deep ensemble learning.


Finally, other methods try to model light reflection properties more explicitly. In \cite{wang2023surface}, light intensity is made dependent on its direction. In \highlight{LightDepth \cite{rodriguez2023lightdepth}, LD for short}\todo{R1.1}, a light decline model, coupled with estimated albedo and shading, is utilised as a supervision signal instead of the standard pose estimation network. They account for non-Lambertian properties by adding a specular loss term. The most closely related method to ours, IID-SfMLearner (IID for short) \cite{li2024image}, uses an intrinsic decomposition network to simultaneously estimate depth, albedo and shading.  To compensate for non-Lambertian properties, they incorporate a shading adjustment network. However, the models described in \cite{li2024image} and \cite{rodriguez2023lightdepth} can only compensate for small light changes and are still not capable of fully handling saturated specularities. In this paper, we improve on \cite{li2024image} by explicitly modelling a non-Lambertian image decomposition (albedo, shading, and specularities) instead of utilising an adjustment network.


\section{Methodology}\label{sec:method}

\subsection{Training}
\subsubsection{Basic Monocular Depth Model}\label{sec:base}
Monocular depth estimation aims at estimating the scene depth of every pixel in a single frame. Self-supervision in monocular depth estimation relies on reconstructing a source image from the viewpoint of a target image. 

Consider source and target images $I_s, I_t$ that visualise the same scene under different viewpoints. These images are fed into networks $\phi_{Depth}$ and $\phi_{Pose}$ that respectively estimate the scene depth maps $D_t, D_s$ and the relative pose $T_{t\to s}$ between $I_t$ and $I_s$. Estimated pose ($T_{t\to s}$), depth ($D_t$), and known camera intrinsics $K$ are used to reconstruct the target from the source image $I_{s\to t}$ following the pixel relation in \Cref{eq:basicrecon}. The supervision signal comes from encouraging the reconstructed image $I_{s\to t}$ to be closer to the target image $I_t$  using a photoconsistency loss (\Cref{eq:basicloss})
, \highlight{such that the weighting factor $\alpha = 0.85$ \cite{godard2019digging,li2024image}}\todo{R1, R2, R3}.

\begin{equation}\label{eq:basicrecon}
    p_s  \approx KT_{t\to s}D_t(p_t)K^{-1}p_t
\end{equation}

\begin{equation}\label{eq:basicloss}
    L_{r}( I_{s \to t}, I_{t}) = \alpha \frac{1 − SSIM( I_{s \to t}, I_{t})}{2}+ (1 − \alpha)\left\| I_{s \to t} − I_{t} \right\|_1
\end{equation}
\subsubsection{IID}

IID \cite{li2024image} (\Cref{fig:compArch}) extends this basic approach with an additional network $\phi_{Decompose}$ that decomposes the source image into albedo ($A$) and shading ($S$). The photometric loss described above is then computed by comparing a reconstructed source image $AS_s$ (instead of $I_s$) against target $I_t$. IID also uses an adjustment network $\phi_{adjust}$ to learn small light offsets.

\subsubsection{Proposed Method}

\begin{figure}[ht!]
    \centering
    \includegraphics[width=\linewidth]{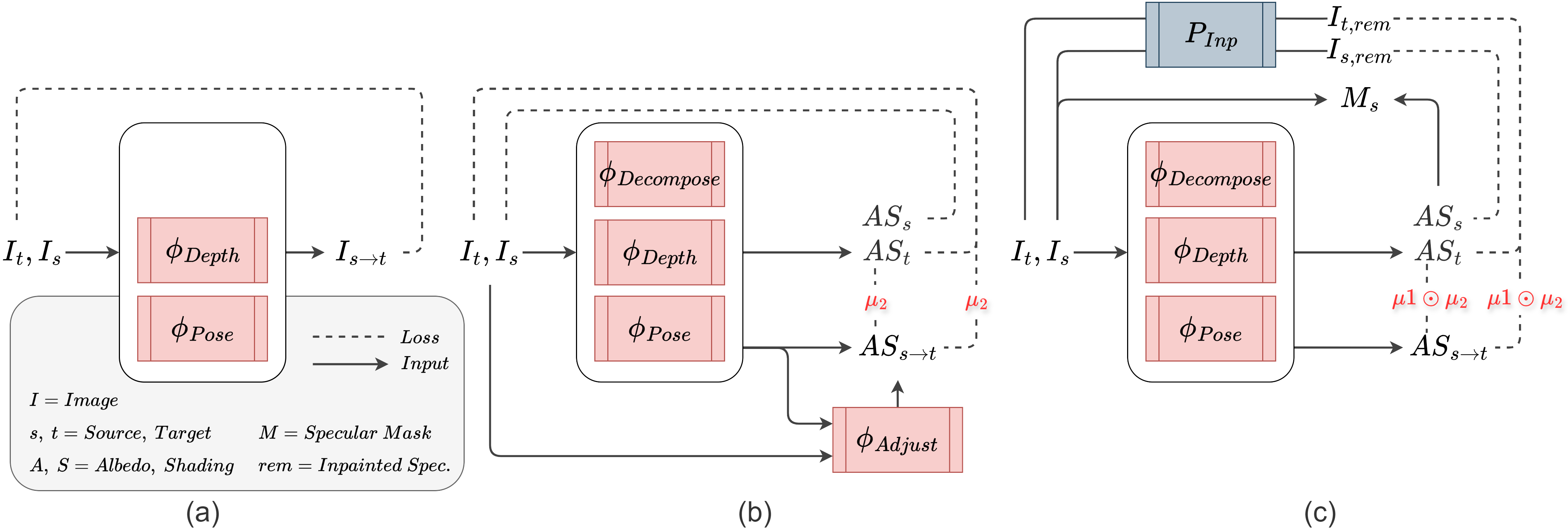}
    \caption{\highlight{A high-level representation of depth estimation training. (a) The basic self-supervision relies on reconstructing a source image from the viewpoint of a target image ($I_{s \rightarrow t}$. (b) The system proposed in \cite{li2024image} (IID) extends the basic approach with Lambertian decomposition ($\Phi_{Decompose} \rightarrow $ I=AS ), auto-masking ($\mu_2$), and a light adjustment network, $\Phi_{Adjust}$. (c) Our proposed system extends IID with non-Lambertian decomposition (I=AS+M) through a pre-trained inpainting network ($P_{Inp}$) and two auto-masking techniques ($\mu_1 \odot \mu_2$) without the need for an adjustment network.}}
    \todo[inline]{R2.2, R3}
    \label{fig:compArch}
\end{figure}

Extending \cite{li2024image} and introducing the insights from \cite{shi2017learning}, we consider a more complete image decomposition (\Cref{fig:compArch}) that includes albedo, shading, and specularities ($M$). For the photometric loss, we compare a reconstructed source image warped to target $AS_{s\to t}$ against an inpainted target with removed specularities ($I_{t,rem}=I_t-M$). The specularity component $M$ is effectively a replacement for IID's offset network, $\phi_{adjust}$, that more explicitly considers that the dominant light changes are specularities.

\begin{figure}[ht!]
    \centering
    \includegraphics[width = \linewidth]{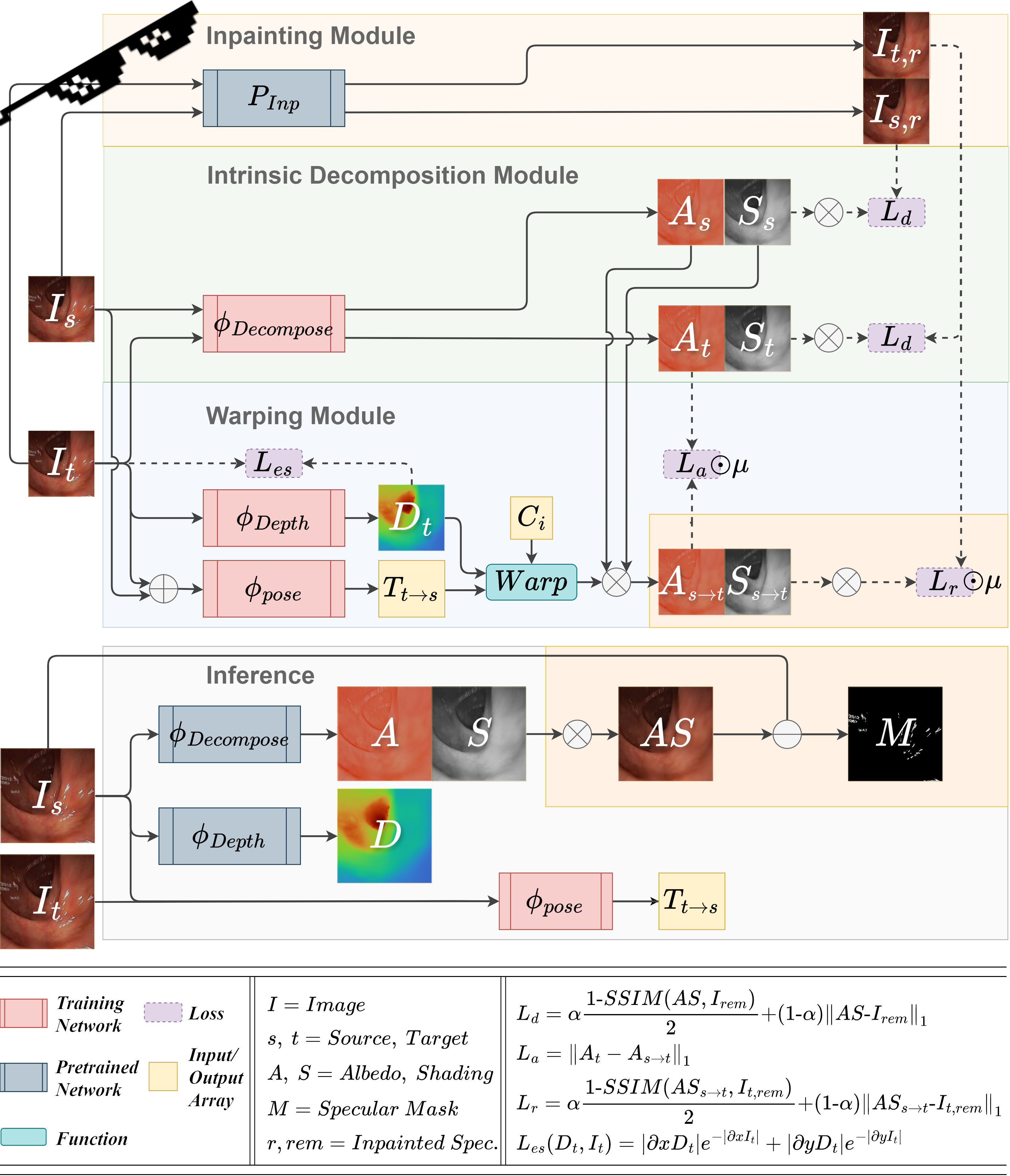}
    \caption{\highlight{Flowchart of the proposed system. During training we compare a reconstructed source image warped to target $AS_{s\to t}$ against an inpainted target with removed specularities ($I_{t,rem}$) through the loss $L_r$, while making sure the depth is smooth ($L_{es}$) and the decomposition is self-supervised through $L_d$ and $L_a$. At inference time albedo, shading, pose, and depth are estimated ($A, S, T, D$) and from those a reconstructed specular free image ($AS$) and a specular mask ($M$) are also generated. Our contributions are highlighted in orange.}}
    \todo[inline]{R2.2, R3}
    \label{fig:flowchart}
\end{figure}

Our complete model
, \highlight{\textit{\textbf{SHADeS}}, which stands for \textbf{SH}ading, \textbf{A}lbedo, \textbf{D}epth and \textbf{S}pecularities}\todo{R1, R2, R3}, 
has the following components: inpainting module, intrinsic decomposition module, warping module, and auto-masking as shown in \Cref{fig:flowchart}.

\noindent
\textbf{Inpainting Module} uses a pre-trained inpainting model $P_{Inp}$ \cite{daher2023temporal}. This model uses a non-learning method \cite{el2011automatic} to segment specularities before inpainting them. The inpainted images $I_{s,rem}, I_{t,rem}$ are used in photoconsistency 
 and deomposition losses.

\noindent
\textbf{The Intrinsic Decomposition Module} uses a U-shaped network, $\phi_{Decompose}$, adopted from \cite{li2024image}, that decomposes the input images into Albedo $A$ and shading $S$ (without specularities). This model is guided by the decomposition loss (\Cref{eq:ld}) making sure the reconstructed image from albedo and shading is similar to $I_{rem}$. Unlike the Lambertian image decomposition assumption ($I = AS$) used in \cite{li2024image} where specular highlights are ignored, we use $I_{rem}$ instead of $I$ since a more accurate non-Lambertian model is $I = AS+ M \implies  AS = I - M  \implies AS \approx I_{rem}$. The albedo loss in \Cref{eq:alb} also guides the intrinsic decomposition model \cite{li2024image}. We apply this loss to ensure that the albedo is influenced solely by warping.

\begin{equation}\label{eq:ld}
    L_{d}( AS, I_{rem}) = \alpha \frac{1 − SSIM( AS, I_{rem})}{2}+ (1 − \alpha)\left\| AS - I_{rem} \right\|_1
\end{equation}

\begin{equation}\label{eq:alb}
    L_{a} =\left\| A_t − A_{s \to t} \right\|_1
\end{equation}

\noindent
\textbf{The Warping Module} consists of pose and depth estimation networks, $\phi_{Depth}$ and $\phi_{Pose}$, following the basic self-supervision strategy described in the first paragraph of \Cref{sec:base}. We introduce the reconstruction loss (\Cref{eq:lr}) adapted from \cite{li2024image} by replacing $I$ with $I_{rem}$. We use their edge-aware smoothness loss to ensure smoothness along the depth gradient (\Cref{eq:es}). We also omit the shading adjustment network $\phi_{Adjust}$ proposed in \cite{li2024image} because it does not impact the results empirically.

\begin{equation}\label{eq:lr}
    L_{r}( AS_{s \to t}, I_{t,rem}) = \alpha \frac{1 − SSIM( AS_{s \to t}, I_{t,rem})}{2}+ (1 − \alpha)\left\| AS_{s \to t} − I_{t,rem} \right\|_1
\end{equation}
\begin{equation}\label{eq:es}
    L_{es}(D_t, I_t) = \left| \partial{x}D_t \right|e^{−\left| \partial{x} I_t \right|} + \left| \partial{y} D_t \right| e^{−\left|  \partial{y} I_t \right|}
\end{equation}

\noindent
\textbf{Two Auto-Masking Techniques} were adopted and applied to the $L_{a}$ and $L_{r}$ losses. The first auto-masking technique of \Cref{eq:am1} from Monodepth2 \cite{godard2019digging} reduces the problem of infinite depth with objects that move with the camera such as overlayed text and shapes from the endoscopic system. The second auto-masking technique from \cite{li2024image} tackles the problem of missing regions between frames due to camera movement (\Cref{eq:am2}). The final mask is their element-wise multiplication $\mu =\mu_1 \odot \mu_2$.

\begin{equation}\label{eq:am1}
    \mu_1 = \underset{s}{min}\  L_{r}(I_t, I_{s\to t}) < \underset{s}{min}\  L_{r}(I_t, I_s )
\end{equation}
\begin{equation}\label{eq:am2}
     \mu_2 = I_{s\to t} > 0
\end{equation}
	
\noindent
\textbf{The Final Loss} in \Cref{eq:finalloss} is composed of the decomposition, albedo, reconstruction, and smoothness losses. Here $\lambda_{d}$, $\lambda_{a}$, $\lambda_{r}$, and $\lambda_{es}$ are set to 0.2, 0.2, 1, and 0.01 as advised in \cite{li2024image}.

\begin{equation}\label{eq:finalloss}
L = \lambda_{d}(L_{d}( AS_t, I_{t,rem}) + L_{d}( AS_s, I_{s,rem})) + \lambda_{a}L_{a} \odot \mu
+ \lambda_{r}L_{r} \odot \mu + \lambda_{es} L_{es}
\end{equation}

\noindent
\textbf{Our Modifications} and the difference between the proposed system training and \cite{li2024image} include the removal of their adjustment module, the incorporation of the inpainting module affecting $L_{d}$ and $L_{r}$, and the addition of the auto-masking technique for static pixels proposed in \cite{godard2019digging}. These modifications are highlighted in orange in \Cref{fig:flowchart} and visually summarized in \Cref{fig:compArch} with a high-level comparison of methods. We also train on real colonoscopy data as opposed to ex-vivo data used in \cite{li2024image}.

\subsection{Inference} 
At inference time, a single frame is used to generate pose, depth, albedo, shading, inpainted image ($AS$), and specularity mask ($M=binarize(I - AS),\  theshold =50$).

\section{Experiments}\label{sec:exp}
\subsection{Data} 
The following datasets were used in our experiments:
\begin{itemize}
    \item $Data_{real}$  --  A colonoscopy dataset from Hyper Kvasir \cite{borgli2020hyperkvasir} with a Boston Bowel Preparation Scale of 2 or 3, which indicates high-quality mucosal views. We used 16,976 frames for training and 786 for testing. A cap of 926 frames per video was set.
    \item $Data_{phantom}$ -- A phantom dataset from C3VD \cite{bobrow2023} with 22 video sequences (10,015 images). This data was used for testing generalizability.
\end{itemize}


\highlight{
All images were first cropped to square and then resized to 288x288. Next, all datasets were undistorted using the camera intrinsics and distortion coefficients from $Data_{phantom}$ \cite{bobrow2023}. These parameters were applied to both $Data_{phantom}$ and $Data_{real}$, as the latter did not provide its own intrinsics, and the $Data_{phantom}$ parameters provided the most reasonable approximation. We observed that applying this undistortion yielded better results than leaving the data uncorrected. This approach aligns with common practices for datasets lacking camera intrinsics, where parameters are estimated when unavailable \cite{godard2019digging}.}\todo{R1.4}

\subsection{Models} 
For comparison, we train $Monodepth2$ \cite{godard2019digging}, $MonoViT$ \cite{zhao2022monovit}, $IID$ \cite{li2024image}, and our method, $SHADeS$, which adds an inpainting module (IM), $\mu_1$ auto-masking (AM), and removes the adjustment network (no adjustment: NA) from $IID$ and thus we also refer to it as $SHADeS_{IM,AM,NA}$ for clarity. To analyze the importance of our modifications, we perform ablation studies by training $IID$ with the added $\mu_1$ auto-masking, $IID_{AM}$. We also train the proposed model without the adjustment network, $SHADeS_{IM,AM}$, and without $\mu_1$ auto-masking, $SHADeS_{IM}$.

\subsection{Setup} 
All experiments were performed on an NVIDIA V100-DGXS. For training, we follow the parameters and implementation of each method. However, we remove flipping since the camera centre is not in the image centre. The number of training epochs was also changed from 30 to 20 for $IID$. For training $SHADeS$, we followed the same parameters of $IID$ with the changes described. 

We initialize $SHADeS$ and $IID$ with their pre-trained depth model \cite{li2024image}. For $Monodepth2$ \cite{godard2019digging} and $MonoViT$ \cite{zhao2022monovit}, we also used their pre-trained models (mono\_640x192) for initialization. However, both $IID$ and $MonoViT$ did not provide a pre-trained model for the pose network, thus we used the Monodepth2's pre-trained pose model to initialize them and $SHADeS$.

\subsection{Evaluations} 
\highlight{We calculate metrics for each image and then compute the average across all images.}\todo{R1.2} The metrics we rely on are:
\begin{enumerate}
    \item \textbf{Specularity Surrounding Metric (SSM):} Since $Data_{real}$ lacks ground truth, we evaluate performance in specular regions segmented using \cite{el2011automatic}. \highlight{Note that this segmentation method was also used during training, introducing a potential bias.}\todo{R1.3} We calculate the percentage of specular regions whose mean depth ($Mean_{spec}$) is close to their surrounding mean depth ($Mean_{surr}$) within a bounding box. This portrays the method's ability to generate smooth depth maps along specularities. More details can be found in the supplementary material.

    \item \highlight{\textbf{Direct Error Metrics following LD \cite{rodriguez2023lightdepth} ($MAE$,
    $MedAE$,
    $RMSE$,
    $RMSE_{log}$,
    $Abs_{Rel}$,
    $Sq_{Rel}$,
    $\delta < 1.25$,
    $\delta < 1.25^2$,
    $\delta < 1.25^3$):}}\todo{R1.2} To evaluate the generalizability of the models, we calculate standard depth metrics between $Data_{phantom}$ and the ground truth. To scale the predicted depth maps, we use median scaling \cite{rodriguez2023lightdepth} where the ratio between the median ground truth and the median prediction is applied.

\end{enumerate}

\section{Results \& Discussion}\label{sec:results}

\begin{figure}[h!]
    \centering
    \includegraphics[width=\linewidth]{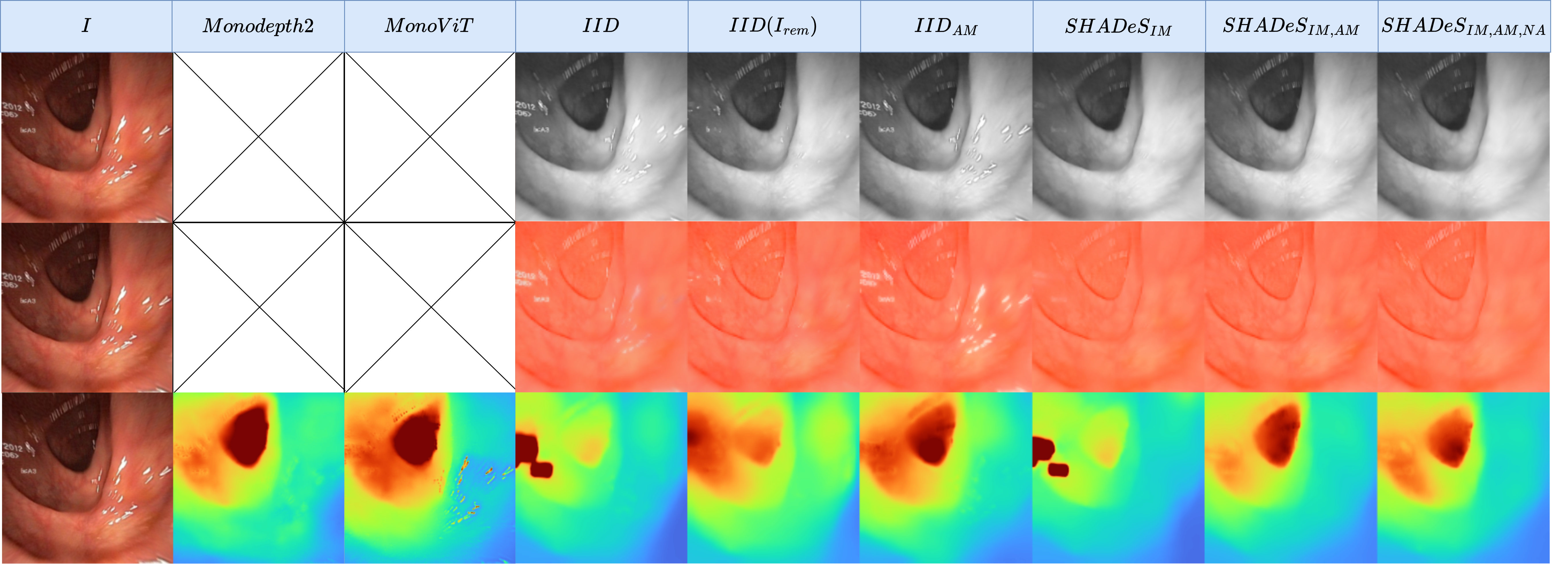}
    \caption{\highlight{Visual results of estimated shading, albedo, and depth on $Data_{real}$. For visual clarity, we clip the depth at 0.8.}}
    \todo[inline]{R2.2, R3}
    \label{fig:resultshk}
\end{figure}

Qualitative and quantitative depth estimation results on $Data_{real}$ are shown in \Cref{fig:resultshk} row 3 and \Cref{tab:resultsquanthk}. From these results, \highlight{we can see that Monodepth2 and MonoViT, which are not tailored for the medical field, perform the worst; This shows the importance of image decomposition. Furthermore, inpainting images as a preprocessing step ($IID_{inp}$) does not positively affect IID showing that one-step solutions without preprocessing can perform as well and even significantly better (e.g. all $SHADeS$ variations)}\todo{R1.2, R1.3, R3}. We also find that methods with an inpainting module (IM) perform better than others, particularly in specular regions. We also notice that $\mu_1$ auto-masking (AM) degrades $IID$ while improving our more realistic non-Lambertian model. We also note the importance of auto-masking in removing the infinite depth effect of static pixels such as text overlays on the image. Finally, removing the adjustment module (NA) did not impact results significantly, which suggests that the adjustment module is unnecessary with our method. 

The same conclusions can be made for albedo and shading when looking at visual results in rows 1 and 2 of \Cref{fig:resultshk} with even more obvious improvements in specular regions. We also notice that the albedo and shading with IM methods are even better than the albedo and shading of $IID(I_{rem})$. This suggests that the model does not only learn to inpaint these specularities but also learns to detect and inpaint specularities not detected by the inpainting pipeline's segmented maps $M_{trad}$ \cite{el2011automatic}.

\begin{table}[h!]
\setlength{\tabcolsep}{2pt}

\centering
\resizebox{\textwidth}{!}{%
\begin{tabular}{r|c c c c c c c c } 
Methods& Monodepth2& MonoViT& $IID$& $IID(I_{rem})$& $IID_{AM}$& $SHADeS_{IM}$& $SHADeS_{IM,AM}$& $SHADeS_{IM,AM,NA}$\\
\hline
SSM (\%)& 39.1& 41.8& 63.7& 62.8& 43.2& 68.3& \textbf{70.6}& \underline{70.0}\\

\end{tabular}}
\caption{\highlight{Specularity Surrounding Metric (SSM) results on $Data_{real}$. SSM evaluates the percentage of smooth depth specular regions by comparing the depth in those regions to their surroundings.}}
\todo[inline]{R1.2, R2.2, R3}
\label{tab:resultsquanthk}
\end{table}

\begin{figure}[h!]
    \centering
    \includegraphics[width=0.5\linewidth]{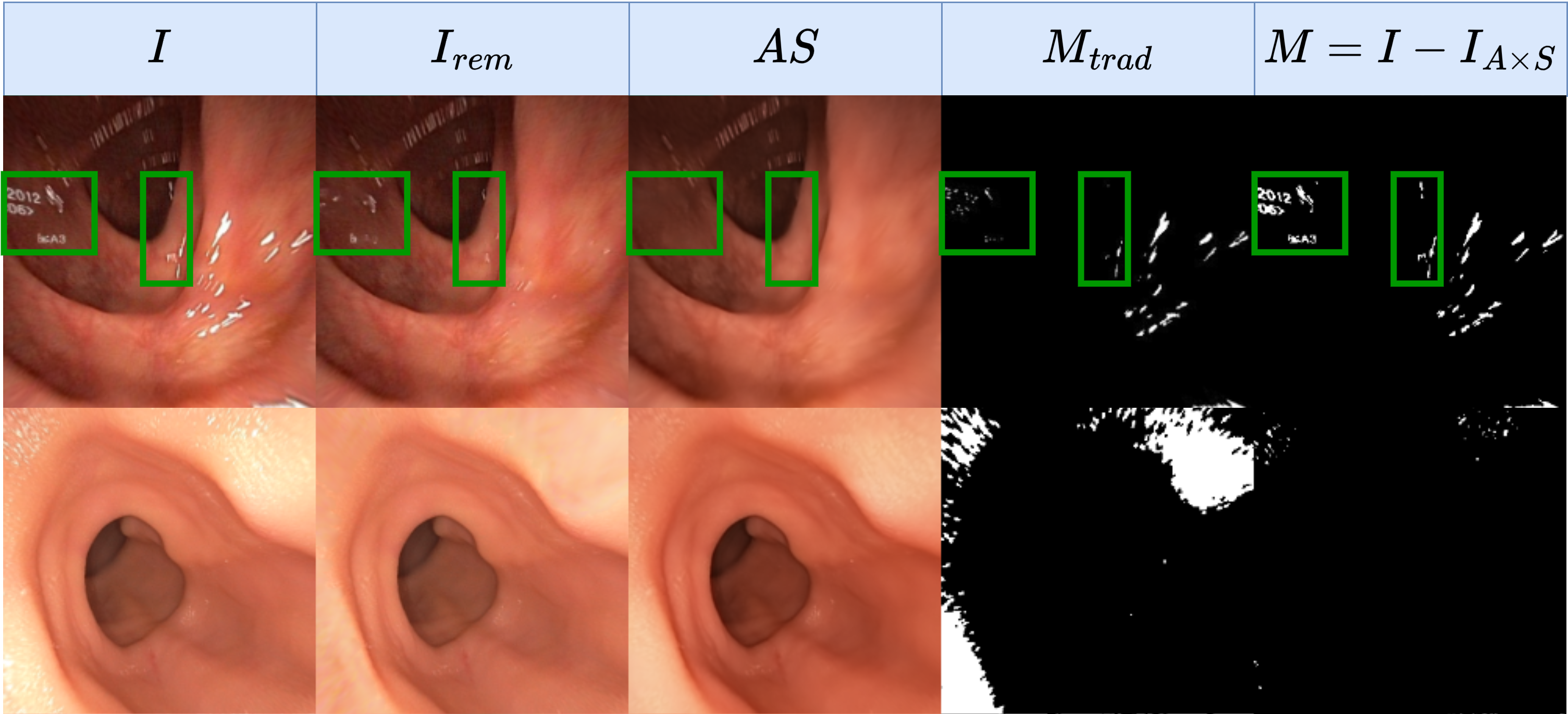}
    \caption{Results on (row 1) $Data_{real}$ and (row 2) $Data_{phantom}$ showing estimated reconstructed images $AS$ and specularity masks $M$ versus their counterparts ($I_{rem}, M_{trad}$) from \cite{daher2023temporal}.}
    \label{fig:resultshkrecon}
\end{figure}

This learnt specularity knowledge can also be seen in the reconstructed image (AS) and specularity mask (M) in row 1 \Cref{fig:resultshkrecon}, where both improved from the traditional methods used within the inpainting module in training \cite{daher2023temporal,el2011automatic}.

\begin{figure}[h!]
    \centering
    \includegraphics[width=\linewidth]{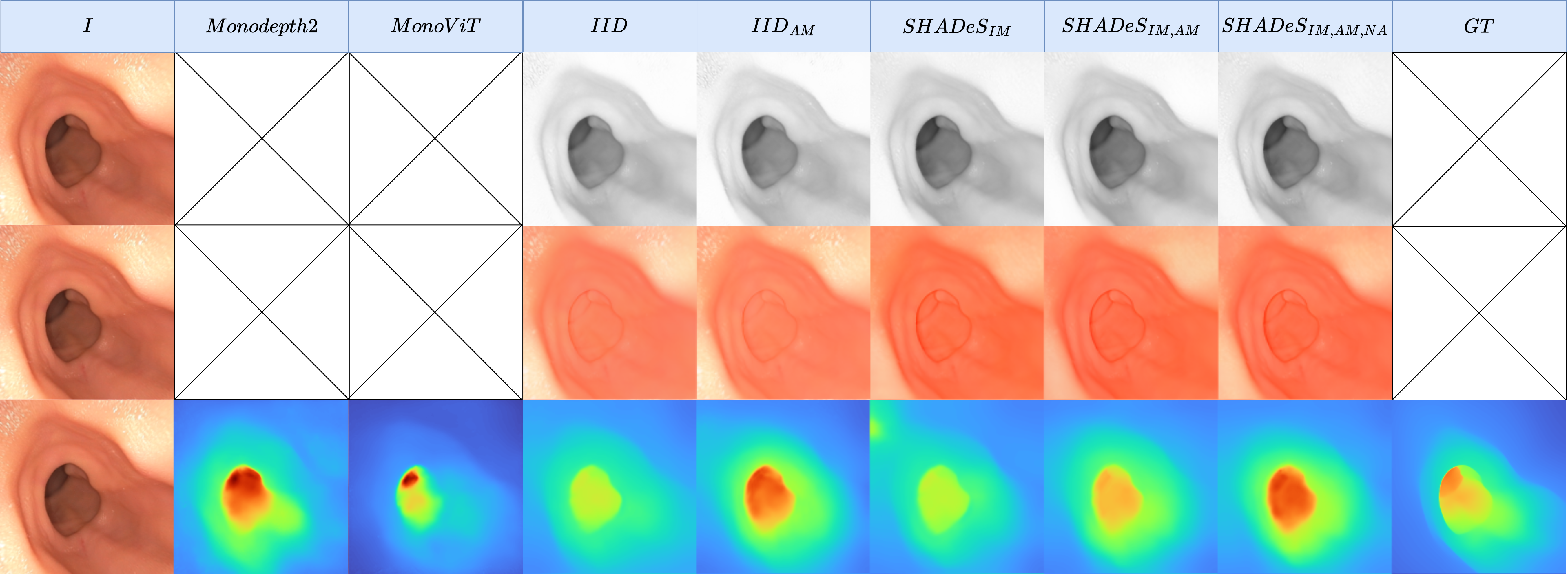}
    \caption{\highlight{Visual results of estimated shading, albedo, and depth on $Data_{phantom}$.}}
\todo[inline]{R1.2, R2.2, R3}
    \label{fig:resultsc3vd}
\end{figure}

\highlight{
To analyze our model's generalizability, we evaluate on $Data_{phantom}$. Quantitatively (\Cref{tab:resultsquantc3vd}), $SHADeS_{IM,AM,NA}$ slightly outperforms other methods. Qualitatively (row 3 \Cref{fig:resultsc3vd}), depth estimates are similar across methods, likely due to all being trained on real data, making generalization to phantom data challenging. In conclusion, our method $SHADeS_{IM,AM,NA}$ generalizes on par with SOTA methods while still improving albedo and shading in specular regions (rows 1, 2 \Cref{fig:resultsc3vd}).}\todo{R2.1, R3}

\begin{table}[h!]
\setlength{\tabcolsep}{2pt}
\centering
\resizebox{\textwidth}{!}{%
\begin{tabular}{r|c c c c c c c c }     
 Methods& Monodepth2& MonoViT& $IID$& $IID(I_{rem})$& $IID_{AM}$& $SHADeS_{IM}$& $SHADeS_{IM,AM}$& $SHADeS_{IM,AM,NA}$\\
 \hline
 $MAE \downarrow$& 4.6 & 5.0 & 4.6 & 4.6 & 4.7 & 4.8 & 4.5 & \textbf{4.4} \\
 $MedAE \downarrow$& 3.3 & 3.1 & 3.2 & 3.3 & 3.3 & 3.4 & \textbf{3.0} & 3.1 \\
 $RMSE \downarrow$& \textbf{6.3} & 7.4 & 6.8 & 6.6 & 6.7 & 7.0 & 6.4 & \textbf{6.3}\\
 $RMSE_{log} \downarrow$& 0.1694 & 0.1667 & 0.1856 & 0.1855 & 0.1714 & 0.1972 & \textbf{0.1607} & 0.1609 \\
 $Abs_{Rel} \downarrow$& 0.1384 & 0.1391 & 0.1476 & 0.1481 & 0.1420 & 0.1590 & 0.1314 & \textbf{0.1312} \\
 $Sq_{Rel} \downarrow$& 1.0198 & 1.2823 & 1.2793 & 1.1903 & 1.1288 & 1.5685 & 0.9879 & \textbf{0.9599} \\
 $\delta < 1.25 \uparrow$& 0.8114 & 0.8230 & 0.8053 & 0.8031 & 0.8162 & 0.7817 & 0.8396 & \textbf{0.8397} \\
 $\delta < 1.25^2 \uparrow$& 0.9839 & 0.9839 & 0.9613 & 0.9612 & 0.9763 & 0.9547 & 0.9855 & \textbf{0.9858} \\
 $\delta < 1.25^3 \uparrow$& 0.9987 & \textbf{0.9991} & 0.9926 & 0.9920 & 0.9971 & 0.9887 & 0.9984 & 0.9985 \\

\end{tabular}}
\caption{\highlight{Depth estimation quantitative results (in mm) on $Data_{phantom}$ with best results in bold.}}
\todo[inline]{R1.2, R2.2, R3}
\label{tab:resultsquantc3vd}

\end{table}

\section{Conclusion}\label{sec:conclusion}
This paper introduces a non-Lambertian self-supervised model that decomposes a single image into its intrinsic components, shading, albedo, depth, and specularity map (SHADeS). Our model improves over Lambertian methods by generating and utilizing an additional specular component. In comparison to state-of-the-art methods, results on real data (Hyper Kvasir) show the robustness of our method to specularities visually and using a specularity smoothness depth metric. Our model can also generalize to phantom data (C3VD) as demonstrated visually and quantitatively (RMSE).

\backmatter

\section*{Acknowledgements}
This research was funded in part, by the Wellcome/EPSRC Centre for Interventional and Surgical Sciences (WEISS) [203145/Z/16/Z]; the Engineering and Physical Sciences Research Council (EPSRC) [EP/P027938/1, EP/R004080/1, EP/P012841/1]; the Royal Academy of Engineering Chair in Emerging Technologies Scheme; H2020 FET (GA863146); and the UCL Centre for Digital Innovation through the Amazon Web Services (AWS) Doctoral Scholarship in Digital Innovation. For the purpose of open access, the author has applied a CC BY public copyright licence to any author accepted manuscript version arising from this submission.

\bibliography{sn-bibliography}

              


\end{document}